\documentclass[10pt,twocolumn,letterpaper]{article}

\usepackage{iccv}
\usepackage{times}
\usepackage{epsfig}
\usepackage{pdfpages}
\usepackage{graphicx}
\graphicspath{{images/}}
\usepackage{amsmath}
\usepackage{amssymb}
\usepackage[utf8]{inputenc}
\usepackage[T1]{fontenc}
\usepackage{textcomp}
\usepackage{multirow}
\usepackage{booktabs}
\usepackage{colortbl}
\usepackage{enumitem}
\usepackage{algorithm}
\usepackage{algpseudocode}
\usepackage{listings}
\usepackage{tablefootnote}

\usepackage[flushleft]{threeparttable}
\definecolor{mygray}{gray}{.9}

\usepackage[breaklinks=true,bookmarks=false]{hyperref}

\iccvfinalcopy 

\def\iccvPaperID{13} 

\ificcvfinal\fi

\begin{document}

\title{Speak2Label: Using Domain Knowledge for Creating a Large Scale Driver Gaze Zone Estimation Dataset}
\author{\parbox{16cm}{\centering
    {\large Shreya Ghosh$^{1}$ Abhinav Dhall$^{1,2}$ Garima Sharma$^{1}$ Sarthak Gupta$^{3}$  Nicu Sebe$^{4}$}\\
    {\normalsize
    $^{1}$Monash University $^{2}$Indian Institute of Technology Ropar $^{3}$Kroop AI $^{4}$University of Trento}\\
    {\small \texttt{ \{shreya.ghosh,abhinav.dhall,garima.sharma1\}@monash.edu sarthak@kroop.ai niculae.sebe@unitn.it}}}
}

\maketitle
\ificcvfinal\thispagestyle{empty}\fi

\begin{abstract}
Labelling of human behavior analysis data is a complex and time consuming task. In this paper, a fully automatic technique for labelling an image based gaze behavior dataset for driver gaze zone estimation is proposed. Domain knowledge is added to the data recording paradigm and later labels are generated in an automatic manner using Speech To Text conversion (STT). In order to remove the noise in the STT process due to different illumination and ethnicity of subjects in our data, the speech frequency and energy are analysed. The resultant Driver Gaze in the Wild (DGW) dataset contains 586 recordings, captured during different times of the day including evenings. The large scale dataset contains 338 subjects with an age range of 18-63 years. As the data is recorded in different lighting conditions, an illumination robust layer is proposed in the Convolutional Neural Network (CNN). The extensive experiments show the variance in the dataset resembling real-world conditions and the effectiveness of the proposed CNN pipeline. The proposed network is also fine-tuned for the eye gaze prediction task, which shows the discriminativeness of the representation learnt by our network on the proposed DGW dataset. Project Page: \url{https://sites.google.com/view/drivergazeprediction/home}
\end{abstract}

\section{Introduction}
One of the primary drivers of progress in deep learning based human behavior analysis is availability of large labelled datasets~\cite{soleymani2014corpus,kellnhofer2019gaze360}. It is observed that the process of labelling becomes non-trivial for complicated tasks. In this paper, we argue that by adding domain knowledge about the task during the data recording paradigm, one can automatically label the dataset quickly. The behavior task chosen in this paper is estimation of driver gaze in car. Distracted driving is one of the main causes of traffic accidents~\cite{gliklich2016texting}. It is important to understand the far-reaching negative impacts of this killer, which is particularly common among younger drivers~\cite{gliklich2016texting,harrispoll}. According to a World Health Organization report~\cite{who}, there were 1.35 million road traffic deaths globally in 2016 and it is increasing day by day. In order to prevent this, efforts are being made to develop \emph{Advanced Driver Assistance Systems (ADAS)}, which will ensure smooth and safe driving by alerting the driver or taking control of the car (handover), when a driver is distracted or fatigued. One important information, which some ADAS needs is a driver's gaze behaviour, in particular, where is the driver looking?
\begin{figure*}[t]
\centering
\includegraphics[width = 1.1 in,height=0.9 in]{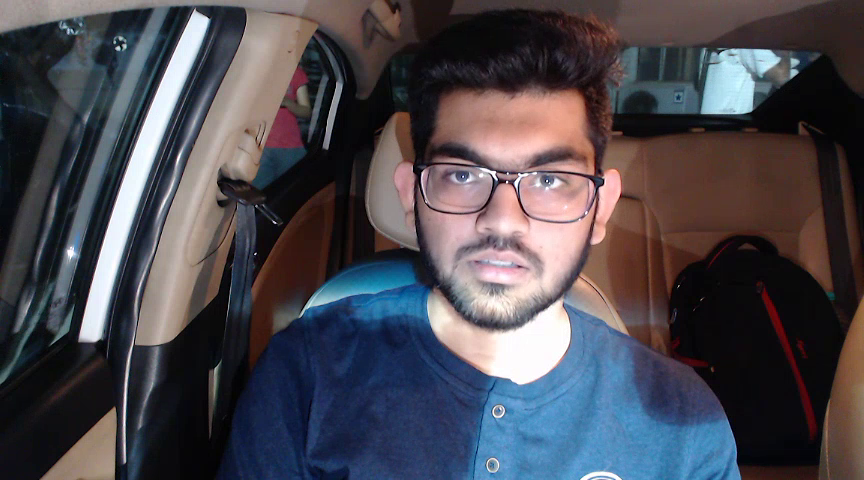}
\includegraphics[width = 1.1 in,height=0.9 in]{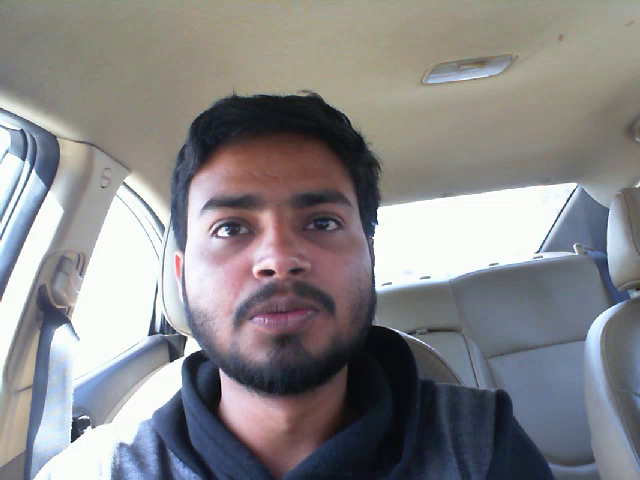}
\includegraphics[width = 1.1 in,height=0.9 in]{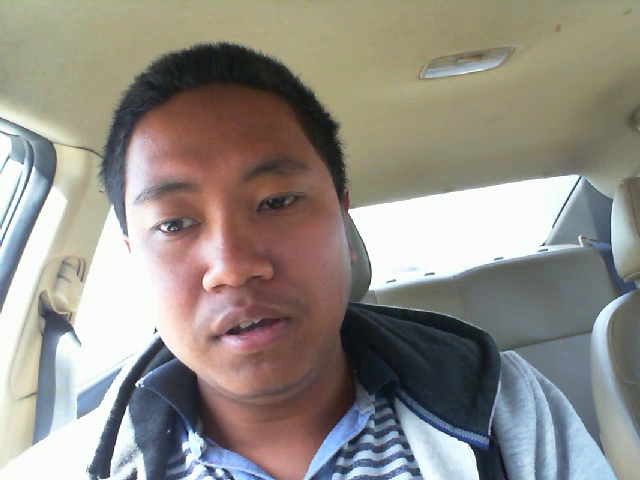}
\includegraphics[width = 1.1 in,height=0.9 in]{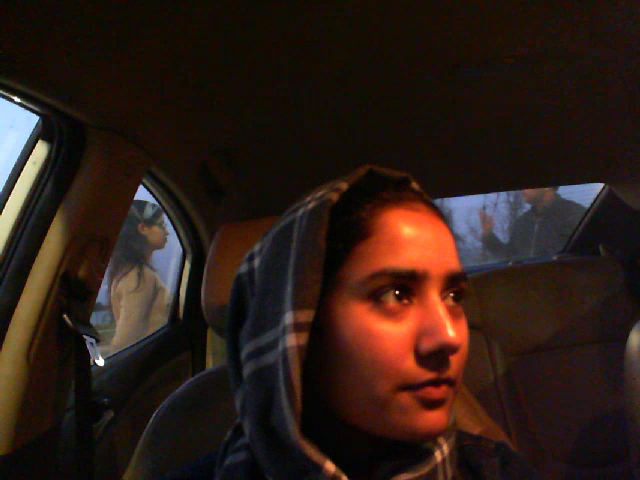}
\includegraphics[width = 1.1 in,height=0.9 in]{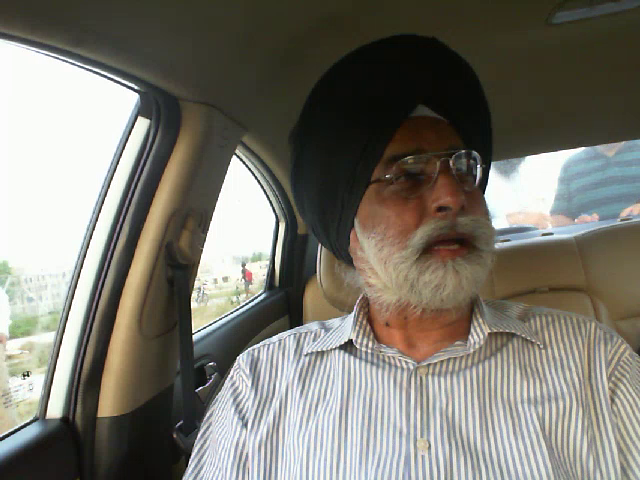} 
\includegraphics[width = 1.1 in,height=0.9 in]{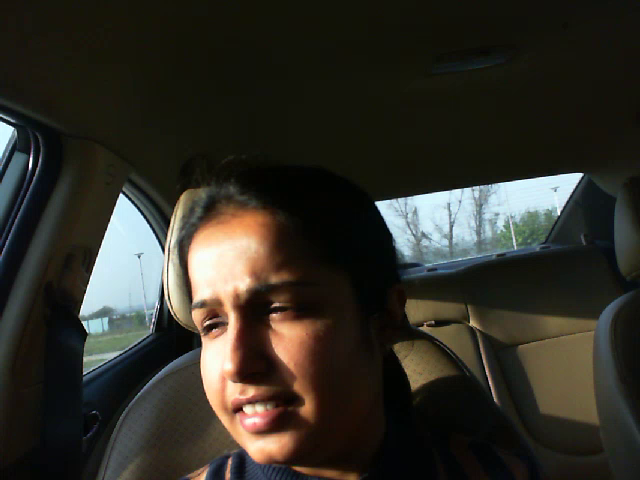}
\vspace{0.1cm}
\caption{\small The proposed Driver Gaze in the Wild (DGW) dataset. Please note the different recording environments and age range.}
\label{fig:sample_images}
\vspace{-5mm}
\end{figure*}
Over past few years, monitoring driver's behaviour as well as visual attention have become interesting topics of research~\cite{xia2007ir,li2013modeling}. Analysis of driver's sparse gaze zone provides an important cue for understanding a driver's mental state. In vision-based driver behaviour monitoring systems, coarse gaze direction prediction instead of exact gaze location is usually acceptable~\cite{jha2018probabilistic,wang2017head,vora2018driver,tawari2014robust,vora2017generalizing,wang2019continuous}. The coarse gaze regions are defined as the in-vehicle areas, where drivers usually look at while driving, for e.g.\ windshield, rear-view mirror, side mirrors, speedometer etc. As per recent studies~\cite{wang2019continuous,jha2018probabilistic}, head pose information is also relevant in predicting the gaze direction. This hypothesis fits well with real and natural driving behaviour. In many cases, a driver may move both head and eyes, while looking at a target zone. Accurate driver's gaze detection requires a very specific set of sensors~\cite{robinson1963method,xia2007ir}, which capture detailed information about the eyes and pupil movements but these can cause an unpleasant user experience. Additionally, manual data labelling is a tedious task, which requires time as well as domain knowledge. In this work, fully automatic labelling can be fairly quickly done by introducing speech during the dataset recording. Our Speech to Text (STT) based labelling technique reduces the above-mentioned limitations of the earlier works. Moreover, this paper can provide help in collaborative driving scenarios, while the vehicle operates in semi-autonomous mode. The \textbf{main contributions} of this paper are as follows:
\begin{itemize}[topsep=1pt,itemsep=0pt,partopsep=1ex,parsep=1ex,leftmargin=*]
    
    \item Traditionally, computer vision based datasets are either labelled manually or using sensors, which makes the labelling process complicated. To the best of our knowledge, this is the first method for speech based automatic labelling of human behavior analysis dataset. The process of labelling in our paper is automatic and does not require any invasive sensors or manual labelling. This makes the task of collecting and labelling data with fairly large number of participants faster. Proposed method requires lesser time ($\sim$30 sec) generating the labels as compared to manual labelling ($\sim$10 min). Additionally, the voice frequency-based detection is used for extracting data samples missed by automatic speech to text method.
    
    \item An `in the wild' dataset - Driver Gaze in the Wild (DGW) containing 586 videos of 338 different subjects is collected. To the best of our knowledge, this is the largest publicly available driver gaze estimation dataset (Fig.~\ref{fig:sample_images}).
    
    \item As the dataset has been recorded with different illumination conditions, a convolutional layer robust to illumination is proposed.
    \item We have performed eye gaze representation learning to judge the generalization performance of our network. (See Supplementary Material)
\end{itemize}


\section{Prior Work}
\label{sec:Prior Work}
\noindent With the progress in autonomous and smart cars, the requirement for automatic driver monitoring has been observed and researchers have been working on this problem for a few years now. For driver's attention estimation, eye tracking is the most evident solution. It is done via sensors or by using computer vision based techniques. Sensor based tracking mainly utilize dedicated sensor integrated hardware devices for monitoring driver's gaze in real-time. These devices require accurate pre-calibration and additionally these devices are expensive. Few examples of these sensors are Infrared (IR) camera~\cite{johns2007monitoring}, contact lenses~\cite{robinson1963method}, head-mounted devices~\cite{jha2018probabilistic,jha2017challenges} and other systems~\cite{feng2013low,zhang2017exploring}. 
All of these above-mentioned systems have sensitivity towards outdoor lighting, difficulty in hardware calibration and system integration. Additionally, constant vibrations and jolts during driving can effect system's performance. 
Thus, it is worthwhile to investigate image processing based zone estimation techniques. 

\begin{table}[t]
\caption{\small Comparison of in-car gaze estimation datasets.} 
\label{tab:DatabseComparison} 
\centering
\begin{tabular}{c|c|c|c|c}
\toprule[0.4mm]
\multicolumn{1}{c|}{\textbf{Ref.}} & \multicolumn{1}{c|}{\textbf{\# Sub}} & \multicolumn{1}{c|}{\textbf{\# Zones}} & \multicolumn{1}{c|}{\textbf{Illumination}} & \multicolumn{1}{c}{\textbf{Labelling}}  \\ \hline \hline

\cite{choi2016real}   & 4                                                              & 8                                                               & \begin{tabular}[c]{@{}c@{}}Bright \&\\ Dim\end{tabular}        & 3D Gyro.                                                      \\ \hline
\begin{tabular}[c]{@{}l@{}}\cite{lee2011real} \end{tabular} & 12                                                             & 18                                                               & Day                                                            & Manual                                                      \\ \hline
\begin{tabular}[c]{@{}l@{}}\cite{fridman2015driver} \end{tabular} & 50                                                             & 6                                                               & Day                                                            & Manual                                                      \\ \hline
\cite{tawari2014driver}                                           & 6                                                              & 8                                                               & Day                                                            & Manual                                                      \\ \hline
\cite{vora2018driver}                                              & 10                                                             & 7                                                               & \begin{tabular}[c]{@{}c@{}}Diff. \\ day times\end{tabular} & Manual                                                      \\ \hline
\cite{jha2018probabilistic}                                                 & 16                                                             & 18                                                              & Day                                                            & \begin{tabular}[c]{@{}c@{}}Head-\\ band\end{tabular} \\ \hline
\cite{wang2019continuous}                                                &                                                           3     & 9                                                               & Day                                                            & \begin{tabular}[c]{@{}c@{}}Motion\\ Sensor\end{tabular}                                                  \\ \hline
\begin{tabular}[l]{@{}l@{}}Ours \end{tabular}                                                     & 338                                                            & 9                                                               & \begin{tabular}[c]{@{}c@{}}Diff. day times\end{tabular}          & Automatic                                                     \\
\bottomrule[0.4mm]
\end{tabular}
\vspace{-5mm}
\end{table}

Prior studies for vision based gaze tracking are mainly focused on two types of zone estimation methods: head-pose based only~\cite{mukherjee2015deep,wang2019continuous} and both head-pose and eye-gaze based~\cite{vasli2016driver,tawari2014driver}. In an interesting work, Lee et al.~\cite{lee2011real} introduced a vision-based real-time gaze zone estimator based on a driver's head moment mainly composed of yaw and pitch. Further, Tawari et al.~\cite{tawari2014robust} presented a distributed camera based framework for gaze zone estimation using head pose only. Additionally,~\cite{tawari2014robust} collected a dataset from naturalistic on-road driving in streets, though containing six subjects only. For the gaze zone ground truth determination, human experts manually labelled the data. Driver's head pose provides partial information regarding the his/her gaze direction as there may be an interplay between eye ball moment and head pose~\cite{fridman2016owl}. Hence, methods totally relying on head pose information may fail to disambiguate between the eye movement with fixed head-pose. Later, Tawari et al.~\cite{tawari2014driver} combined head pose with horizontal and vertical eye gaze for robust estimation of driver's gaze zone. Experimental protocols are evaluated on the dataset collected by~\cite{tawari2014robust} and it shows improved performance overhead moment~\cite{tawari2014robust}. In another interesting work, Fridman et al.~\cite{fridman2015driver,fridman2016owl} proposed a generalized gaze zone estimation using the random forest classifier. They validated the methods on a dataset containing 40 drivers and with cross driver testing (test on the unseen drivers). When the ratio of the classifier prediction having the highest probability to the second highest probability is greater than a particular threshold, the decision tree branch is pruned.
Similarly,~\cite{vasli2016driver} combined 3D head pose with both 3D and 2D gaze information to predict gaze zone via a support vector machine classifier. Choi et al.~\cite{choi2016real} proposed the use of deep learning based techniques to predict categorized driver's gaze zone. 
Recently, Wang et al.~\cite{wang2019continuous} proposed an Iterative Closet Points (ICP) based head pose tracking method for appearance-based gaze estimation. 
The labelling is performed initially using a head motion sensor and later clustering is used on this head pose technique. The labels in this case, do not consider the scenario, where there is a difference between the eye gaze and the head pose of a subject. In another interesting work, Jha et al.~\cite{jha2018probabilistic} map 6D head pose (three head position and three head rotation angles) to an output map of quantized gaze zone. The users in their study wear headbands, which are used to label the data using the head pose information only. Few of the selected methods are also described in Table~\ref{tab:DatabseComparison}. Please note that most of the datasets are not available publicly, with the exception of Lee et al.~\cite{lee2011real}, though it contains 12 subjects only. It is easily observable that our proposed dataset DGW has a large number of subjects and more diverse illumination settings. Further, the methods discussed above require either manual labelling of the driver dataset or it is based on a wearable sensor. We argue that the labelling of gaze can be noisy and erroneous task for labellers due to the task being monotonous. Further, with wearable sensors such a headband, it may be uncomfortable for some subjects. Therefore, in this work, we propose an alternate method of using speech as part of the data recording. This removes the need for manual labelling and the user having to wear any headgear as well. 
Similarly, we are interested in predicting the zone, where the driver is looking at? This considers the both eye gaze, head pose (Fig.~\ref{fig:headpose}) and the interplay of gaze and head pose. Nowadays, self-supervised and unsupervised learning is getting attention as it has the potential to overcome the limitation of supervised learning based algorithms, which requires large amount of labelled data. A few recent works~\cite{leo2014unsupervised,lu2016estimating, liu2019exploiting,huang2020learning,yu2019unsupervised,yu2019improving} explored this domain. The labelling technique used by us in this work also exploits the domain knowledge (speech) and helps in labelling a fairly large dataset quickly.

\section{DGW Dataset}
\label{sec:Dataset}
We curate a new driver gaze zone estimation dataset as the datasets in this domain are small in size and are mostly not available for academic purpose. Fig.~\ref{fig:sample_images} shows the frames from the proposed DGW dataset. Please note that the dataset and the baseline scripts will be made publicly available. 

\begin{figure}[b] 
\centering 
\includegraphics[width = 0.7 in]{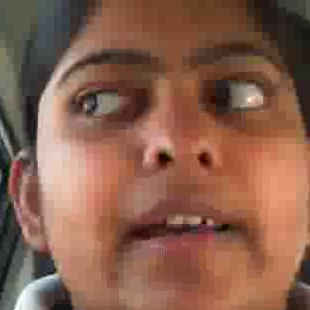}
\includegraphics[width = 0.7 in]{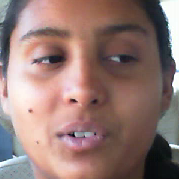}
\includegraphics[width = 0.7 in]{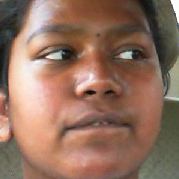}
\includegraphics[width = 0.7 in]{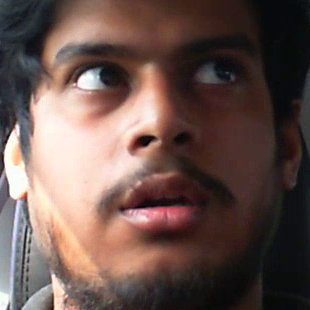}
\caption{\small Challenging samples from our DGW dataset in which the head pose and eye gaze differ for subjects. The labels should not just be based on the head pose as in prior works~\cite{jha2018probabilistic, wang2019continuous}.}
\label{fig:headpose}
\vspace{-5mm}
\end{figure}

\begin{figure}[t]
\centering
\includegraphics[width = \linewidth]{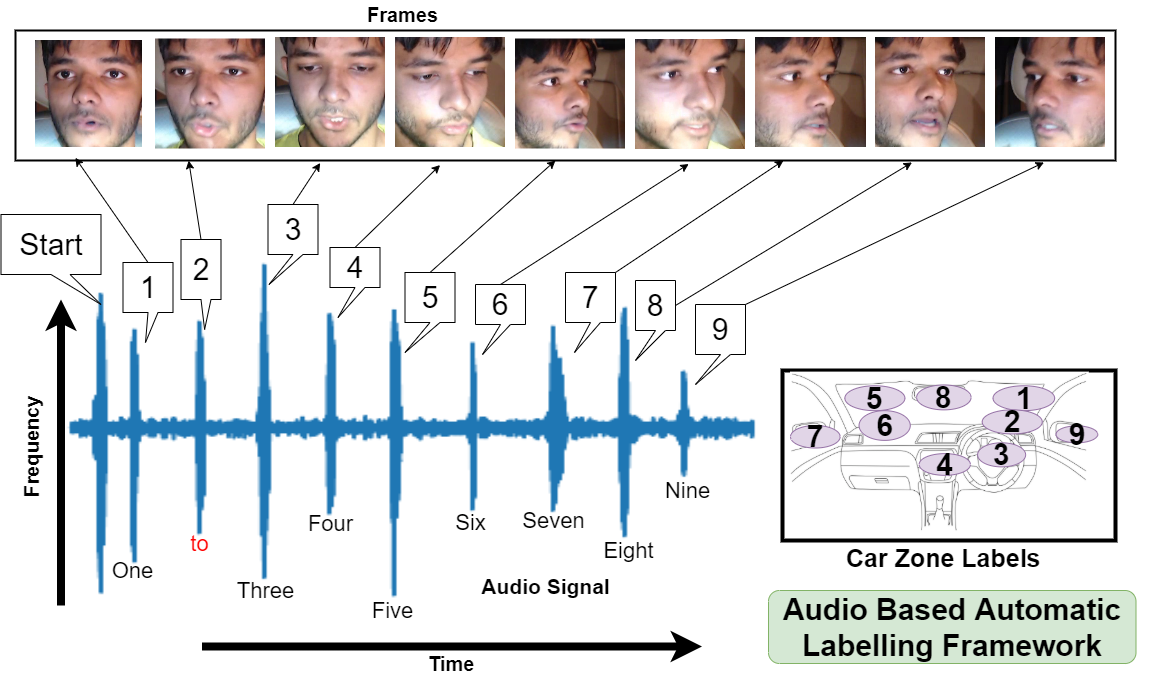}
\caption{\small Overview of the automatic data annotation technique. On the top are the representative frames from each zone. Please note the numbers written in alphabets below the curve. The red colored `to' shows the incorrect detection by the STT library. This is correct by frequency analysis approach in Sec.~\ref{sec:RepairLabel}. On the bottom right are reference car zones.}
\label{fig:carregions}
\vspace{-3mm}
\end{figure}

\noindent \textbf{Data Recording Paradigm.}
Before data collection, consent was taken from participants regarding the scope of data usage. This included agreement to share data with university and industrial labs and if a face of a participant could be used in any publication in the future. We pasted number stickers on different gaze zones of the car (Fig.~\ref{fig:carregions}). The nine car zones are chosen from back mirror, side mirrors, radio, speedometer and windshield. The recording sensor used is a Microsoft Lifecam RGB. For recording the following protocol is followed: 
We asked the subjects to look at the zones marked with numbers in different orders. For each zone, the subject has to fixate on a particular zone number and speak the zone's number and then move to the next zone. For recording realistic behaviour, no constraint is mentioned to the subjects about looking by eye movements and/or head movements. The subjects choose the way in which they are comfortable. This leads to more naturalistic data (see Fig.~\ref{fig:headpose}).
For the subjects who wear spectacles, if it is comfortable for the participant, they are requested to record twice i.e. with and without the spectacles.
The RA was also present in the car and observed the subject and checked the recorded video. If there was a mismatch between the zone and the gaze, the subject repeated the recording. This insures correct driver gaze to car zone mapping. 

The data is collected during different times of the day for recording different illumination settings (as evident in Fig.~\ref{fig:sample_images}). Recording sessions are also conducted during the evening after sunset at different locations in the university. This enables different sources of illumination from street lights (light emitting diodes, compact fluorescent lamp and sodium vapour lamps) and also from inside the car. There are a few sessions during which the weather was cloudy. This brings healthy amount of variation in the data. 

\section{Automatic Data Annotation}\label{sec:RepairLabel}
As manual data labelling can be an erroneous and monotonous task, our method is based on automatic data labelling. Following are the details of the labelling process.

\noindent \textbf{Speech To Text.}
Post extraction of the audio from the recorded samples, the IBM Watson's STT API~\cite{stt} is used to convert the audio signal into text. We searched for the keywords `one', `two', `three', `four', `five', `six', `seven', `eight' and `nine' in the extracted speech in ascending order. As we recorded the data in `one' to `nine' in sequence, therefore, sequentially ordered texts having high probability is considered. Further, we extracted the frames corresponding to the detected time stamps by adding an offset (10 frames chosen empirically) before and after the detection of the zone number. We used the US English model (16 kHz and 8 kHz). In a few cases, this model was unable to detect correctly, this may be due to different pronunciation of English words across different cultures. In order to overcome this limitation, we applied STT rectification.  

\noindent \textbf{STT Rectification.}
Generally, human voice frequency lies in the range of 300-3000 Hz~\cite{titze1998principles}. We use the frequency and energy domain analysis of the audio signal to detect time duration of the audio signal from the data. The assumption based on the recording paradigm of DGW dataset is that the numbers are spoken in a sequence. If during scan of the numbers generated from the STT process, there is a mismatch for a particular number, the following steps are executed to find the particular zone's data: \textit{Step 1:} Convert stereo input signal to mono audio signal and start scanning with a fixed window size $T$. Calculate frequency over the time domain of the audio for a window size ~$T$. \textit{Step 2:} If the frequency lies in the human voice range 300-3000 Hz, then this window is a probable candidate. \textit{Step 3:} Compute ratio between energy of speech band and total energy for this window. If ratio is above a threshold, then this window is a probable candidate. \textit{Step 4:} If there is an overlap between the timestamps generated from steps 3 and 4 above, the zone label is assigned to the frames between the timestamps. This process extracted an extra 4000 frames, which were earlier missed due to the noise generated from STT. We checked manually for some recordings randomly, most of the useful data has been extracted following the steps above. \textit{Refer supplementary document for dataset statistics and validation of automatic data annotation process.} Please note that the label generated after STT is treated as ground truth label.

\subsection{Label Refining} \label{LabelRefine}
 Our proposed automatic data annotation may generate noisy labels during the gaze transition between two zones in the car. For example, a subject utters the word `one' and looks at region `one'. After that the subject shifts gaze from region `one' to region `two' and utters `two'. During the transition between these two utterances, some frames may have been incorrectly annotated. Similarly, in a few cases, the zone utterance and the shifting of gaze may not have occurred simultaneously. To handle such situations, we perform label rectification based on an auto-encoder network followed by latent features based clustering. 


\noindent \textbf{Encoder-Decoder.}
The encoder part of the network is based on the backbone network (Inception-V1, refer Fig.~\ref{fig:network}). The decoder network consists of series of alternate convolution and up-sampling layers. The details of decoder network is as follows: the convolution layers have 1024, 128, 128, 64 and 3 kernels having $3\times3$ dimension. The first up-sampling layer has $2\times2$ kernel. The second and third up-sampling layers have $4\times4$ kernel. It is to be noted that the facial embedding representation learnt in this network encodes the eye gaze with the head pose information.

\noindent \textbf{Clustering.} After learning the auto-encoder, we perform k-means clustering on the facial embeddings. Here the value of k=9 is same as the number of zones. After k-mean clustering of all the samples, previous labels of the transition frames are updated on the basis of its Euclidean distance from the cluster center. More specifically, we measure the distance between a transition frame's facial embedding with the 9 cluster centers and assign the frame the label of the nearest neighbor cluster. Please note that it is a static process. The refined labels are then considered as the ground truth label for the dataset.  

\section{Method}
\label{sec:Method}



\begin{figure*}[t]
\centering
\includegraphics[width = \linewidth,height = 32mm]{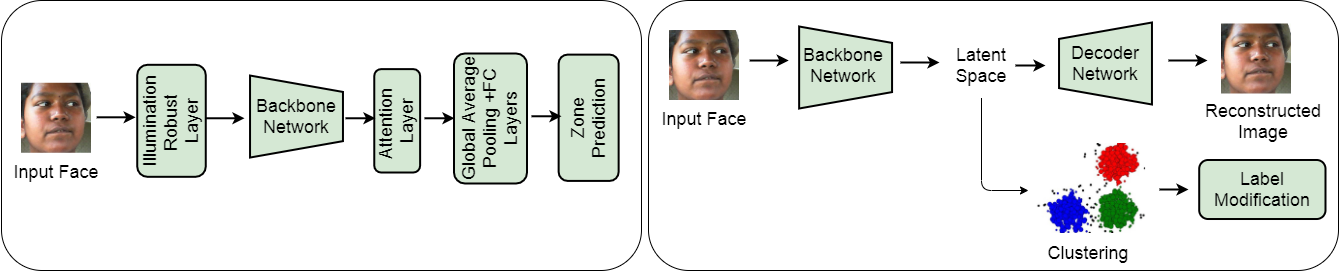} 
\caption{\small LEFT: Overview of the proposed network. RIGHT: Label refinement architecture.}
\label{fig:network}
\vspace{-3mm}
\end{figure*}

\noindent \textbf{Baseline.}
For the baseline methods, we experiment with several standard networks like Alexnet~\cite{krizhevsky2012imagenet}, Resnet~\cite{he2016deep} and Inception Network~\cite{szegedy2016rethinking}. The input to the network is the cropped face computed using the Dlib face detection library. The proposed network is shown in Fig.~\ref{fig:network}. The baseline network takes $224\times224\times3$ facial image as input. From the results using the standard networks mentioned above, one of the limitations observed is that as the DGW dataset has been recorded in diverse illumination conditions, some samples, which contained illumination change across the face are mis-classified. Sample images can be seen in Fig.~\ref{fig:fail+illu} top. To the backbone network, we add the illumination layer presented below.

\noindent \textbf{Illumination Robust Layer.} 
For illumination robust facial image generation, we follow a common assumption proposed by Lambert and Phong~\cite{phong1975illumination}. The authors adopted the concept of ideal matte surface, which obey Lambert's cosine law. The law states that the incoming incident light at any point of an object surface is diffused uniformly in all possible directions. Later, Phong has added a specular highlight modelization term with Lambertian model. This term is independent of the object's geometric shape. Moreover, it is also independent of the lighting direction of each surface point. For illumination robust learning, we follow the computationally efficient Chromaticity property (Zhang et al.~\cite{zhang2019improving}). c = \{r, g, b\} is from the following skin color formation equation:

\vspace{-7mm}
\begin{equation}
\small c_i = \dfrac{f_i {\lambda_i}^{-5} S(\lambda_i)}{{(\prod_{j=1}^{3}f_j {\lambda_j}^{-5} S(\lambda_j))}^{ \dfrac{1}{3}}} \times \dfrac{e^{-\dfrac{k_2}{\lambda_i T}}}{e^{\dfrac{1}{3} \sum_{j=1}^{3}-\dfrac{k_2}{\lambda_j T}}}
\label{eq:skin}
\end{equation}

Here, i = \{1, 2, 3\}, correspond to the R, G \& B channels, respectively. $f_i$ is from Dirac delta function; $\lambda_i$'s are tri-chromatic wavelengths (the wavelengths of {R, G, B} lights wherein \{$\lambda_1 \in [620, 750], \lambda_2 \in [495, 570], \lambda_3 \in [450, 495],$ unit : nm\}); S($\lambda$) is spectral reflectance function of skin surface; $k2 = \dfrac{hc}{k_B}$ ( h: Plank's constant $h = 6.626\times10^{-34}J.s $, $k_B$: Boltzmann's constant $k_B = 1.381\times10^{-23}J.k^{-1}$ and $c = 3 \times 10^8 m · s^{-1}$) refer to first and second radiation constants in Wien's approximation and T represents the lighting color temperature. If we write the Equation (\ref{eq:skin}) in $c_i = A \times B$ format, then the left part of Equation 3 is the illumination robust (A) and right part (B) is illumination dependent due to the colour temperature factor $T$, which varies throughout the dataset. Thus, for illumination robust feature extraction, we initialize a constant kernel having the $T$ independent value of  part (B). $T$ is initialized with a Gaussian distribution. Further, the product of constant and Gaussian kernel is considered for learning.



\noindent \textbf{Attention based Gaze Prediction.}
The eye region of a person's face is important in estimating driver's gaze zone as it gives vital information about the eye gaze. We already show that there are images in DGW dataset (Fig.~\ref{fig:headpose}), where the head pose is frontal even though the driver may be looking at a particular zone, which is not in the front using the change in the eye gaze. Motivated by this hypothesis, we add attention augmented convolution module~\cite{bello2019attention} to the network. Let's consider a convolution layer having $F_{in}$ input filters, $F_{out}$ output filters and $k$ kernels. $H$ and $W$ represent the height and width of an activation map. $d_v$ and $d_k$ denote the depth of values and the depth of queries/keys in MultiHead-Attention (MHA). $v = \dfrac{d_v}{F_{out}}$ is the ratio of attention channels to number of output filters and $ k_a = \dfrac{d_k}{F_{out}}$ is the ratio of key depth to number of output filters.

The \textbf{A}ttention \textbf{A}ugmented \textbf{Conv}olution (\textbf{AAConv})~\cite{bello2019attention} can be written as follows: $AAConv(X) = Concat[Conv(X), MHA(X)]$
Where, X is the input. MHA consists of a $1\times1$ convolution with $F_{in}$ input filters and $(2d_k+d_v) = F_{out}(2k_a+v)$ output filters to compute queries/keys and values. An additional $1\times1$ convolution with $d_v = \dfrac{F_{out}}{v}$ input and output filters is also added to mix the contribution of different key heads. AAConv is robust to translation and different input resolution dimensions. 

\noindent \textbf{Network Architecture.} 
The proposed network architecture is shown in the  left box of Fig.~\ref{fig:network}. In this part of the network, we basically perform the gaze zone classification task with Inception-V1 as backbone network. The input of this network is facial image. The illumination robust layer and attention layer are introduced in the beginning and end of the backbone network to enhance the performance. After attention layer, the resultant embedding is passed through two dense layers (1024, 512) before predicting the gaze zone.




\section{Experiments}
\label{sec:Experimets}


\noindent \textbf{Data partition.} The dataset is divided into train, validation and test sets. The partition is performed randomly. 203 subjects are used in training partition, 83 subjects are used in validation partition and rest of the 52 subjects are used in test partition. Having unique identities in the data partitions helps in learning more generic representations.


\noindent \textbf{Experimental Setup.}
The following experiments were evaluated and compared to understand the complexities of the data and create baselines:
1) Baseline: based on Inception-V1 as the backbone network;
2) Baseline + Illumination Layer: On top of Inception-V1, an illumination robust layer is added;
3) Baseline + Attention: Attention augmented convolution layer is introduced in Inception-V1;
4) Baseline + Illumination Layer + Attention: This is the combination of illumination robust layer and attention;
5) Performance with standard backbone networks: Comparison of several state-of-the-art networks is performed;
6) Ablation study for illumination robust layer's configuration;
7) Eye-gaze representation learning: Transfer learning experiments to check the effectiveness of the representation learnt from DGW; 
8) Evaluation on Nvidia Jetson Nano platform.

\noindent \textbf{Evaluation Matrix and Training Details.}
Overall accuracy in \% is used as evaluation matrix for gaze zone prediction. For gaze representation learning, the angular error (in \textdegree) is used as evaluation matrix for the CAVE dataset~\cite{smith2013gaze} and mean error (in cm) is used for the TabletGaze dataset~\cite{huang2015tabletgaze}. For CAVE the angular error is calculated as $mean\ error \pm std.\ deviation$ (in \textdegree).

\begin{table}[t]
\caption{\small Comparison of backbone networks on the proposed DGW dataset (validation set) with original labels (9 classes).}
\label{tab:base+iilu}
\centering
\scalebox{1}{
\begin{tabular}{l|c|c}
\toprule[0.4mm]
\multirow{2}{*}{\textbf{Networks}} & \multicolumn{2}{c}{\textbf{Accuracy (\%)}} \\ \cline{2-3} 
 & \textbf{Network} & \textbf{\begin{tabular}[c]{@{}c@{}}Network +\\  Illumination Robust Layer\end{tabular}} \\ \hline \hline
Alexnet & 56.25 &  57.98  \\ \hline
Resnet-18 & 59.14 & 60.87  \\ \hline
Resnet-50 & 58.52   & 60.05 \\ \hline
Inception-V1 & 60.10   & 61.46   \\ 
\bottomrule[0.4mm]
\end{tabular}}
\vspace{-6mm}
\end{table}


\noindent For the backbone network, Inception-V1 network architecture is used. For training the following parameters are used: 1) SGD optimizer with 0.01 learning rate with $1 \times e^{6}$ decay per epoch. 2) Kernels are initialized with Gaussian distribution with initial bias value 0.2. 3) In each case, the models are trained for 200 epochs with batch size 32. 

\noindent For gaze representation learning, we fine tuned the proposed network with the following changes. Two FC layers (256, 256 node dense layers for the TabletGaze and 1024, 512 node dense layers for the CAVE dataset) are added with ReLU activation to fine tune the network. The learning rate was set to 0.001 with SGD optimizer. For both these datasets, we froze the first 50 layers of the network and fine tuned the rest part.

\begin{table}[b]
\caption{\small Comparison of the proposed network architecture with and without illumination robust layer and attention. Here, Base: Baseline, Illu: Illumination Layer and Attn:Attention.}
\label{tab:results_all}
\centering
\begin{tabular}{l|l|c|c}
\toprule[0.4mm]
\multicolumn{2}{l|}{\multirow{2}{*}\textbf{\begin{tabular}[c]{@{}c@{}}Inception-V1\\ (Trained with original labels)\end{tabular}}}     & \multicolumn{2}{l}{\textbf{Accuracy (\%)}}                                 \\ \cline{3-4} 
\multicolumn{2}{l|}{}                                           & \multicolumn{1}{l|}{\textbf{Validation}} & \multicolumn{1}{l}{\textbf{Test}} \\ \hline \hline
\multicolumn{2}{l|}{Baseline (7 classes)}                         & 66.56                                    & 67.39                              \\ \hline
\multirow{4}{*}{(9 classes)} & Base                          & 60.10                                    & 60.98                              \\ \cline{2-4} 
                          & Base + Attn                & 60.75                                    & 60.08                              \\ \cline{2-4} 
                          & Base + Illu             & 61.46                                    & 60.42                              \\ \cline{2-4} 
                          & Base + Attn + Illu & 64.46                                    & 62.90                              \\ 
\bottomrule[0.4mm]
\end{tabular}
\vspace{-6mm}
\end{table}

\section{Results}
\label{sec:results}
\subsection{Network Performance}

\noindent \textbf{Experiment with state-of-the-art backbone networks.}
We experimented with several network architectures to get an overview of the trade-off between the number of parameters and accuracy. Specifically, we choose lightweight networks like AlexNet~\cite{krizhevsky2012imagenet}, ResNet-18~\cite{he2016deep}, ResNet-50 and Inception~\cite{szegedy2016rethinking}. Among these networks due to robust handling of different scales, the inception network performs better. The further results are based on the Inception-V1 as the backbone network. Based on this empirical analysis, the baseline network is the Inception-V1 network plus global average pooling and Fully Connected (FC) layers. The quantitative analysis of these networks are shown in Table~\ref{tab:base+iilu}. It also reflects that the addition of illumination robust layer increases the performance. It is effective in real-world scenarios in which different sources of illumination play vital role and many existing techniques may not perform properly. The classification performance increases as illumination robust layer is added to the baseline network as compared to the baseline network only.

\noindent \textbf{9 zones vs 7 zones.} 
Additionally, we experimented with a simpler task i.e. seven gaze zone classification. Although the data is collected with nine zones, zones 1 and 2 can be merged to represent the right half of the windscreen and zones 5 and zone 6 can be merged to represent the left half of the windscreen. We call this experiment setting as `7-zone'. Please refer to Fig.~\ref{fig:carregions} (on Page 4) for car zone label reference. From Table~\ref{tab:results_all}, it is observed that the classification accuracy is higher in the case of 7 classes. The validation and test accuracies increase by 6.46\% and 6.41\%, respectively. This also means that for small sized cars fine-grained zone classification is non-trivial. Please note that all the following experiments in the paper are performed for 9 car zones only. 
\begin{table}[b]
\caption{\small Variation in network performance (in \%) w.r.t the illumination layer size and position.}
\label{tab:illu_ablation}
\centering
\begin{tabular}{l|c|c|c}
\toprule[0.4mm]
\multirow{2}{*}{\begin{tabular}[c]{@{}c@{}}\textbf{Illumination} \\\textbf{Layer}\end{tabular}} & \multicolumn{1}{l|}{\multirow{2}{*}{\textbf{Layer Details}}} & \multicolumn{2}{l}{\textbf{Accuracy (\%)}}                                 \\ \cline{3-4} 
                                             & \multicolumn{1}{l|}{}                                        & \multicolumn{1}{l|}{\textbf{Validation}} & \multicolumn{1}{l}{\textbf{Test}} \\ \hline \hline
\multirow{2}{*}{Dense Layer}                 & 1024                                                         & 56.16                                    & 57.93                              \\ \cline{2-4} 
                                             & 4096                                                         & 58.51                                    & 61.18                              \\ \hline
\multirow{4}{*}{\begin{tabular}[c]{@{}c@{}}Convolution \\ Layer\end{tabular}}           & 32                                                           & 53.48                                    & 52.16                              \\ \cline{2-4} 
                                             & 64                                                           & 60.47                                    & 58.38                              \\ \cline{2-4} 
                                             & \textbf{128}                                                 & \textbf{61.46}                           & \textbf{60.42}                     \\ \cline{2-4} 
                                             & 256                                                          & 57.71                                    & 57.35                              \\ 
\bottomrule[0.4mm]
\end{tabular}
\vspace{-6mm}
\end{table}

\noindent \textbf{Improvement over baseline.}

\noindent \textit{Quantitative Analysis.} Table~\ref{tab:results_all} shows the gradual improvement over the baseline due to the addition of attention and illumination layers. By adding the attention layer, we introduce guided learning. In the next step, we added an illumination robust layer to encode illumination robust features, which also increase the performance of the model. Our final model has both illumination and attention layer followed by FC layers.

\noindent \textit{Significance Test.} One way ANOVA test is performed on the models is to calculate the statistical significance of the models. The p-values of the `Baseline + Illumination', `Baseline + Attention' and `Baseline + Illumination + Attention' models are 0.03, 0.04 and 0.01, respectively. The p-values of the models are $<0.05$, which indicates that the results are statistically significant.

\noindent \textit{Qualitative Analysis.} Fig.~\ref{fig:fail+illu}a shows few examples, where previously mis-classified images are classified correctly after the addition of the illumination layer. We noted that for few subjects with spectacle glare, performance increased.

\noindent \textbf{Label Rectification Performance.} \label{LabelRefine}
To avoid error in automatic labelling process, clustering based label modification (Sec.~\ref{LabelRefine}) was also performed.  

\noindent \textit{Qualitative Analysis.} After clustering, the labels of approximately 400 frames changed. Zone 9 class set changed most with frames in this zone increasing by 5.2\%. Frames in Zone 8 and 5 increased by 3.5\% and  2.3\%. Other zones have less than 1\% increment. This suggests that whenever there is a significant distance change in among consecutive zones, the error increases.

\noindent \textit{Quantitative Analysis.} After label modification, the validation accuracy changes from 64.46\% to 66.44\%  as shown in Table~\ref{tab:result_method}. This supports the hypothesis that the classification accuracy increases for frames in between transition from one zone to another and specially for the zones, with large physical distance (eg: zone 7 and 8).

\noindent \textbf{Test Set Results.} For all of the methods, test set performances was calculated. Both the `Baseline + Illumination' and `baseline+attention' perform slightly lower than the baseline (Table~\ref{tab:results_all}). The combined effect of illumination and attention improves the test set performance from 60.98\% to 62.90\%. Error in automatic labelling could be the cause for this performance. Further, training is performed on `train+validation' set and performance is evaluated on test set. The test performance increased to 64.31\% over the baseline (60.98\%) and `Baseline + Illumination + Attention' network (62.90\%).

\begin{table}[b]
\caption{\small Results of the proposed methods. The `Inception-V1 + Illumination + Attention' model is used for the experiments.}
\label{tab:result_method}
\centering
\begin{tabular}{c|c|c|c|c}
\toprule[0.4mm]
\multirow{2}{*}{\textbf{\begin{tabular}[c]{@{}c@{}}Methods\end{tabular}}} & \multicolumn{2}{l|}{\textbf{Accuracy (in \%)}}                                 & \multicolumn{2}{c}{\textbf{F1 Score}} \\ \cline{2-5} 
                                                                                                                           & \multicolumn{1}{l|}{\textbf{Validation}} & \multicolumn{1}{l|}{\textbf{Test}} & \textbf{Validation}   & \textbf{Test}  \\ \hline \hline
\begin{tabular}[c]{@{}c@{}}Proposed \\ Network\end{tabular}                                                                & 64.46                                    & 62.90                              & 0.52                  & 0.52           \\ \hline
\begin{tabular}[c]{@{}c@{}}Train on \\ (Train + Val)\end{tabular}                                                          & -                                        & 64.31                              & -                     & 0.59           \\ \hline
\begin{tabular}[c]{@{}c@{}}Label \\Modified   \end{tabular}                                                                                                          & 65.97                                    & 61.98                              & 0.63                  & 0.59           \\ 
\bottomrule[0.4mm]
\end{tabular}
\vspace{-5mm}
\end{table}

\noindent \textbf{Results with State-of-the-art Methods.}
We evaluate our method (Sec.~\ref{sec:Method}) on the LISA Gaze Dataset v1~\cite{vora2018driver}, which contains 7 zones. Our method achieves 93.45\% classification accuracy.~\cite{vora2018driver}’s method gives 91.66\% on their own data. This validates the discriminative ability of our proposed network. Further, we evaluate the method proposed by Vora et al.~\cite{vora2018driver} on our data as well. The method achieves 67.31\% and 68.12\% classification accuracy on the validation and test sets, respectively. We evaluate standard networks~\cite{DBLP:journals/corr/HeZRS15,DBLP:journals/corr/SzegedyVISW15,DBLP:journals/corr/SimonyanZ14a,iandola2016squeezenet} and other state-of-the-art methods~\cite{vora2017generalizing,vora2018driver,vasli2016driver,tawari2014driver,fridman2015driver,yoon2019driver} on DGW dataset in Table~\ref{tab:sota_dgw}. It is observed that Resnet 152~\cite{DBLP:journals/corr/HeZRS15} and Inception V3~\cite{DBLP:journals/corr/SzegedyVISW15} perform better than the others, however, the performance is not high as observed for other dataset such as~\cite{vora2018driver}. This can be attributed to the large number of subjects (338) and different illumination conditions under which DGW was recorded.

\begin{table}[b]
\caption{ {Performance comparison with existing CNN-based driver gaze estimation models.}}
\label{tab:sota_dgw}
\centering
\begin{tabular}{l|c|c}
\toprule[0.4mm]
\textbf{ {Method}}                                                                                  & \textbf{ {Val Acc (\%)}} & \textbf{ {Test Acc (\%)}} \\ \hline \hline
VGG 16~\cite{DBLP:journals/corr/SimonyanZ14a}                                                                                  &  {58.67}                                                                           &  {58.90}                                                                    \\ \hline
Inception V3~\cite{DBLP:journals/corr/SzegedyVISW15}                                                                           &  {67.93}                                                                           &  {68.04}                                                                    \\ \hline
Squeezenet~\cite{iandola2016squeezenet}                                                                              &  {59.53}                                                                           &  {59.18 }                                                                    \\ \hline
Resnet 152~\cite{DBLP:journals/corr/HeZRS15}                                                                              &  {68.94}                                                                           &  {69.01 }                                                                   \\ \hline
Vora et al.~\cite{vora2018driver}                                                          &  { 67.31}                                                       &  {68.12}                                                \\ \hline
\begin{tabular}[l]{@{}l@{}}Vora et al.\\(Alexnet face)~\cite{vora2017generalizing}\end{tabular} &  {56.25  }                                                                         &  {57.98 }                                                                   \\ \hline
\begin{tabular}[l]{@{}l@{}}Vora et al. \\(VGG face)~\cite{vora2017generalizing}\end{tabular}     &  {58.67  }                                                                        &  {58.90 }                                                                   \\ \hline
Vasli et al.~\cite{vasli2016driver}    &   {52.60}                                                                        &   {50.41}                                                                   \\ \hline
Tawari et al.~\cite{tawari2014driver}    &   {51.30}                                                                        &   {50.90}                                                                   \\ \hline
Fridman et al.~\cite{fridman2015driver}    &   {53.10}                                                                        &   {52.87}                                                                   \\ \hline
\begin{tabular}[l]{@{}l@{}}Yoon et al.\\(Face + Eyes)~\cite{yoon2019driver}\end{tabular}    &   {70.94}                                                                        &   {71.20}                                                                   \\ \hline
\begin{tabular}[l]{@{}l@{}}Lyu et al.~\cite{lyu2020extract}\\\end{tabular}     &   {85.40}                                                                        &   {81.51}                                                                   \\ \hline
\begin{tabular}[l]{@{}l@{}}Stappen et al.~\cite{stappen2020x}\\\end{tabular}     &   {71.03}                                                                        &   {71.28}                                                                   \\ \hline
\begin{tabular}[l]{@{}l@{}}Yu et al.~\cite{yu2020multi} \\\end{tabular}     &   {80.29}                                                                        &   {82.52}                                                                   \\ 
\bottomrule[0.4mm]
\end{tabular}
\vspace{-5mm}
\end{table}

\subsection{Ablation Study} \label{sec:ablation}

\noindent \textbf{Effect of Illumination layer.}
We also conducted experiments (Table~\ref{tab:illu_ablation})for observing the variation in network performance w.r.t the illumination layer variation.

\noindent \textit{1) Position Vs Performance.}
First, we experiment to check the ideal position of the illumination robust layer. For this analysis, the layer is implemented in the beginning and end conv-layers. For the beginning conv-layer, the performance increased. On the other hand, for the end conv-layer (before the flatten layer) performance did not increase. The reason could be that after several convolutions and max-pooling operation, the information is changed enough to be useful with the illumination robust layer.

\noindent \textit{2) Filter size Vs Performance.} This analysis is conducted in two settings: 1) The robustness is implemented in convolutional layer and 2) The layer is implemented in dense layer (fully connected layer). From the Table~\ref{tab:illu_ablation}, we can observe that as the illumination layer filter size increases, the performance also increases. We used the baseline + illumination + attention framework to compute these results. 

\begin{figure}[t]
    \centering
    \includegraphics[width=\linewidth]{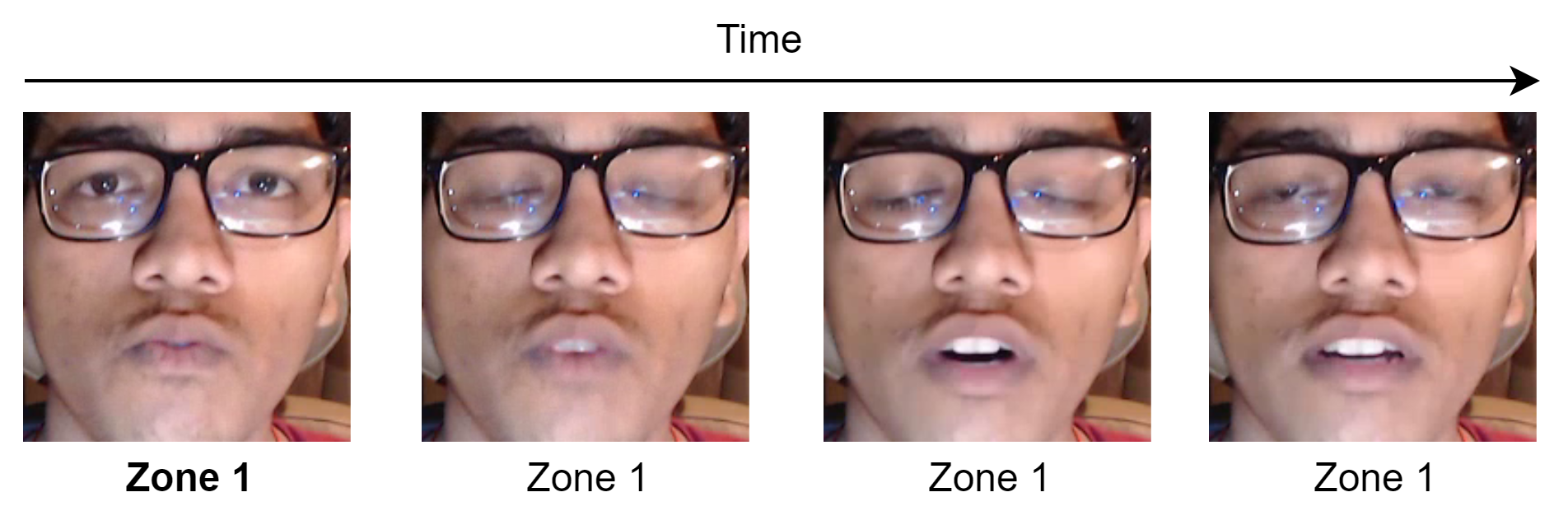}
    \caption{Gaze label assignment during an eye blink. The assigned labels (last three frames) are mentioned below each frame.}
    \label{fig:temporal}
    \vspace{-7mm}
\end{figure}

\begin{figure}[t]
\centering
\includegraphics[width = 0.75 in]{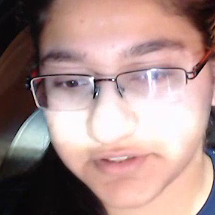}\hspace{0.1mm}
\includegraphics[width = 0.75 in]{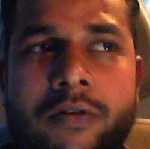}\hspace{0.1mm}
\includegraphics[width = 0.75 in]{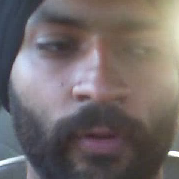}\hspace{0.1mm}
\includegraphics[width = 0.75 in]{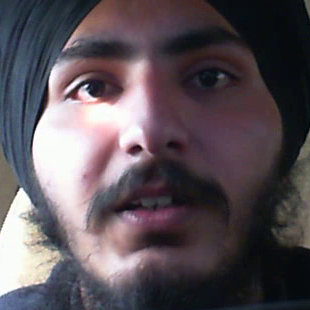}\\
\vspace{1mm}
\includegraphics[width = 0.75 in]{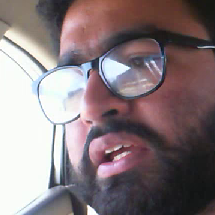}\hspace{0.1mm}
\includegraphics[width = 0.75 in]{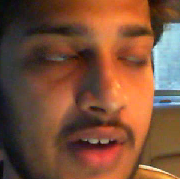}\hspace{0.1mm} 
\includegraphics[width = 0.75 in]{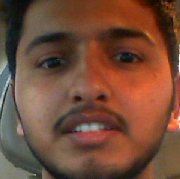}\hspace{0.1mm}
\includegraphics[width = 0.75 in]{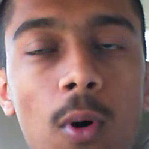}
\caption{\small TOP: Correctly classified samples with illumination robust layer, which were earlier mis-classified by the baseline network. Bottom: Incorrectly classified samples by our network.}  
\label{fig:fail+illu}
\vspace{-8mm}
\end{figure}

\noindent \textbf{Qualitative Analysis of Failure Cases.} Fig.~\ref{fig:fail+illu} shows few mis-classified examples. The reason for this could be incomplete information of the eyes in these samples. All the cases (except bottom left in Fig.~\ref{fig:fail+illu}) eyes are not visible properly and illumination layer even unable to recover information. In these cases head pose and neighbour frame's gaze information could be vital clue to predict gaze zone. Additionally, there could be an interplay between pose and gaze. 

\noindent \textbf{Discussion on Effect of Eye Blink.}
Eye blinks are involuntary and a periodic event. During an eye blink event, a major cue for gaze estimation (i.e. pupil region) is missing. In the absence of the eye information (partially or fully closed eyes), the head pose may still provide useful cues required for gaze estimation. The same is also observed in some gaze datasets (example: Gaze360) i.e. the head pose information is considered as the gaze information in case of partial or complete occlusion scenarios. In our work, we too assume that head pose and eye information provide complementary information. In case of an eye blink, head pose information can be useful. To analyse this in the driver gaze context, we conduct following experiments. If we remove eye blink frames (detected using eye aspect ratio~\cite{rayheadpose}) the validation accuracy improved by 6.27\%, which means that partial eye close or fully closed samples are challenging. However, if we consider practical deployment scenario during which the driver gaze detection system will be used in a car, temporal information in the form of previous frames will also be available. So, there is a possibility of borrowing information from earlier frames, when the current frame has incomplete information due to an eye blink. A simple method is using labels of the neighbour previous frames (where eyes are open) for assigning them to frames containing eye blink. With this assignment, we note that the validation accuracy improves by 4.7\%. This small experiment is an indication that in the presence of an eye blink, we can still consider the information from the previous frames, which leads to correct prediction of current gaze zone. An example is shown in the Fig.~\ref{fig:temporal}, here, the neighbour frame is the first frame in which the eyes are open. We assign the same labels for the subsequent frames (i.e. eye blink frames).

\noindent \textbf{Other.} Please refer the supplementary material for the ablation study regarding gaze representation, effect of lip movement and deployment on Jetson Nano environment.

\section{Conclusion, Limitations and Future Work}
\label{sec:Conclusion, Limitations and Future Work}

In this paper, we show that automatic labelling can be performed by adding domain knowledge during the data recording process. We propose a large scale gaze zone estimation dataset, which is labelled fully automatically using the STT conversion. It is observed that the missed information from STT can be recovered by analyzing the frequency and energy of the audio signal. The dataset recordings are performed in different illumination conditions, which makes the dataset closer to the realistic scenarios. To take care of the varying illumination across the face, we propose an illumination robust layer in our network. The results show that the illumination robust layer is able to correctly classify some samples, which have different or low illumination. Further, the experiments on eye gaze prediction using the features learnt from our network on the DGW dataset show that the features learnt are effective for gaze estimation task. In order to record even more realistic data, car driving also needs to be added in the data recording paradigm. The trickier part is about how to use speech effectively in this case as the drivers will be concentrating on the driving activity. Perhaps, a smaller subset of driving dataset can be labelled using a network trained on the existing stationary recorded dataset and it can be validated by human labellers. Further, the DGW dataset will be extended with more female subjects to balance the current gender distribution. At this point, our method does not consider the temporal information. It will be interesting to understand the effect of temporal information on the gaze estimation as a continuous regression-based problem. Few prior works used IR cameras~\cite{johns2007monitoring, wang2019continuous} for their superior performance in dealing with illumination effects such as on the driver glasses. Our use of a webcam-based RGB camera validated the process of STT labelling and illumination invariance. It will be of interest to try distillation based knowledge transfer from our DGW dataset into the smaller sized network later fine-tuned on smaller gaze estimation datasets. 

Currently, on Nvidia Jetson Nano, we achieve 10 FPS. It should further improve if network optimization techniques such as quantization and separable kernels are experimented with. Our proposed method is implicitly learning the discriminativeness, due to the head pose. In future, we plan to integrate the head pose information explicitly to evaluate its usefulness. We will also evaluate the performance of the network and the usefulness of the learnt features for the task of distracted driver detection. One future direction can be joint gaze zone and distraction detection as a multi-task learning problem. 

{\small
\bibliographystyle{ieee_fullname}
\bibliography{driver_gaze}
}


\section*{Supplementary material}

# Supplementary Material

## 1. Dataset Statistics

Following are the relevant data statistics:

1. **Age:** The curated DFW dataset has variation in subject's age. The age ranges from 18-63 years. Table 1 depicts the group-wise age distribution.

2. **Gender Distribution:** The data consists of 247 male and 91 female subjects.

3. **Daytime Distribution:** 55.2% of the data is collected in daylight and 44.8% during evening with different light sources. The daytime recording sessions were conducted during both sunny and cloudy weathers. The recording was performed in different places in the university resulting in variation due to different illumination sources. Additionally, few recordings have performed at night with very little light source. As face detector failed to detect faces in these videos, these recordings are discarded from the data.

4. **Duration:** The average length of a subject's recording session is 15-20 sec. The data was recorded over one and half month time span. **5) Face Size:** The average face size in the dataset is $179 \times 179$ (in pixel). To measure the face size, Dlib face detection library [7] is used.

5. **Other Attributes:** Specular reflection add more challenge to any gaze estimation data. In DGW data, 30.17% of the subjects have prescribed spectacles. The subjects having prescribed spectacles are requested to record data with and without spectacle (subject to the condition if they can). Thus, two different settings corresponding to these subjects were recorded. Additionally, there are variations in head pose due to sitting posture of the participants.

## 2. Validation of Automatic Data Annotation

**Head Pose Variation.** We analyse the variation in head pose values w.r.t. their zones by computing the density based clustering [3] on the head pose information [4] (i.e. yaw, pitch and roll). We observed that for zones 1-3, the head-pose is mainly centrally concentrated (forming 2, 3 and 2 clusters). For the remaining zones, there are more than 5 clusters each. This experiment indicates that it may be noisy, if the gaze data is based on head-pose only. Fig. 1 shows the clustering results of zones 2, 3, 5 and 7.

**Comparison With Manual Annotation Process.** For comparing the automatic annotation with manual annotation, expert and non-expert annotators are assigned. There were 3 annotators (2 experts and 1 non-expert[1]) who were assigned for this task. We asked the annotators to label 15 videos. Further, we calculate few statistics to judge the quality of labelling. 2 experts take approximately 10-15 min (on average) to annotate each video. The mean squared errors of automatic label with the 2 expert annotator's labels for 15 videos are 0.49 and 0.54 respectively. Similarly, the mean squared error in case of non-expert annotator is 0.78. The cohen's kappa between the expert annotators is 0.8. At microsecond level, there is a high probability of wrong labelling during annotation by human labellers.

## 3. Illumination Robust Layer

As per [11], both Lambertian and Phong models can be formulated by the following equations:

$$L_{diffuse} = S_d E_d \left( n.l \right) \quad (1)$$

$$L_{diffuse} + L_{specular} = S_d E_d \left( n.l \right) + S_s E_s \left( v.r \right)^\gamma \quad (2)$$

In Lambertian Equation (Eq. 1), $S_d$ is diffusion reflection coefficient; $E_d$ denotes the diffuse lighting intensity; $n$ corresponds to normal vector and $l$ denotes normal vector along the direction of incoming light [11]. Similarly, in Phong Equation (Eq. 2), $S_s$ is the specular reflection coefficient; $E_s$ denotes the specular lighting intensity; $v$ is the normal vector along observation direction and $r$ is the normal vector along the reflected light. $\gamma$ is a constant termed as 'shininess constant'.

## 4. Data Pre-processing

After labelling the curated data, few pre-processing methods are performed to remove noise from the training data.
**1) Face Detection.** Dlib face detector [8] is computed with low threshold value as there are large illumination variations in the dataset. As a result of the low threshold value, we observed that in few cases, the false face detection rate also increased.
**2) Optical Flow-based Face Pruning.** In order to deal with the false face detections, we compute dense optical flow [1] across the detected face frames. If two consecutive frames have high Forbenius norm of the optical flow magnitude (i.e. above an empirically decided threshold), we discarded the later one. An example of face pruning is shown in Fig. 2 in which the third frame is discarded due to its high Forbenius norm value. The comparison pairs are also marked in the figure. This removes the incorrectly detected faces in the training set.

---

[1]Expert refers to labeler with prior labelling experience.

Table 1. Age distribution in the DGW dataset.

| Age Range | 18-25 | 26-35 | 36-45 | Over 45 |
|---|---|---|---|---|
| Subjects (in %) | 61.8 | 28.7 | 6.7 | 2.8 |

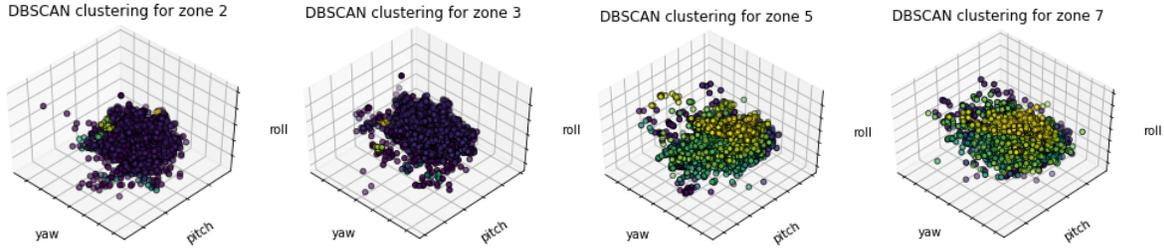

Figure 1. DBSCAN clustering of head pose along yaw, pitch and roll axis for zones 2, 3, 5 and 7 (left to right) respectively. Large number of clusters within a zone, implies that we cannot only rely on head pose information for labelling the data. (best viewed in colour)

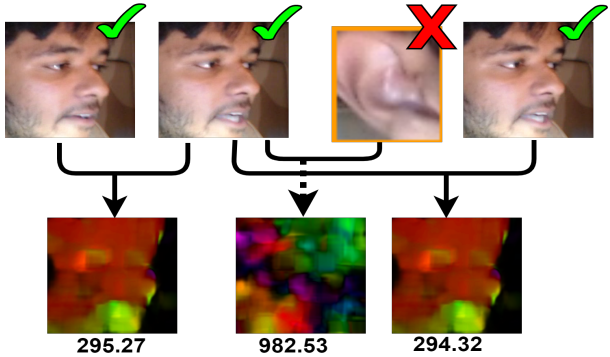

Figure 2. Overview of the optical flow based pruning method. The rejected frame is marked with a red cross.

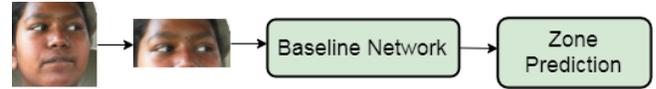

Figure 3. Lip movement effect analysis pipeline.

## 5. Ablation Study

**Eye Gaze Representation Learning.** In order to analyze whether our network learnt a generalized face representation, we extracted the features from weights trained on DGW and fine tuned for the task of eye gaze estimation. We fine-tuned the network on the Columbia gaze [10] (CAVE) and TabletGaze [5] datasets. The results are shown in Table 3 and Table 2, respectively. In the case of CAVE, our model stabilizes the standard deviation. Similarly, for TabletGaze, the fine tuned network works well. These quantitative results indicate that our proposed network has learnt efficient representation from the DGW dataset.

**Effect of Lip Movement.** To check the effect of lips movement on the network, we performed the following experiment: images were cropped from the eye region up to nose tip point and the gaze zone prediction is re-trained. This resulted in drop of overall accuracy for the baseline network. Fig. 3 shows the overall pipeline of the aforementioned experiment. It is interesting to note that the performance difference is minimal due to the effect of loss of head pose information, when the eyes are used as input only.

**Jetson Nano Experiments.** It is important to note that a more complex backbone network may achieve better performance. We chose Inception-V1 due to relatively better performance and to be able to evaluate the performance on a small platform such as the Nvidia Jetson Nano. These networks will be expected to run on close to real-time in a car based computer. On Nvidia Jetson Nano the best network runs at 10 Frames Per Second (FPS) with 34,532,961 parameters.

Table 2. Results on Tablet Gaze with comparison to baselines [5]. Effectiveness of the learnt features from DGW dataset is demonstrated here. TG: TabletGaze, RP: Raw Pixels.

| Methods TG | RP [5] | LoG [5] | LBP [5] | HoG [5] | mHoG [5] | Ours |
|---|---|---|---|---|---|---|
| k-NN | 9.26 | 6.45 | 6.29 | 3.73 | 3.69 | 3.77 |
| RF | 7.2 | 4.76 | 4.99 | 3.29 | 3.17 | |
| GPR | 7.38 | 6.04 | 5.83 | 4.07 | 4.11 | |
| SVR | - | - | - | - | 4.07 | |

Table 3. Results on the CAVE dataset (of °0 yaw angle) using angular deviation, calculated as $mean\ error \pm std.\ deviation$ (in°).

| Dataset | [6] | | [9] | | [2] | | Ours | |
|---|---|---|---|---|---|---|---|---|
| | x | y | x | y | x | y | x | y |
| CAVE | $1.67 \pm 1.19$ | $3.47 \pm 3.99$ | $2.65 \pm 3.96$ | $4.02 \pm 5.82$ | $1.67 \pm 1.19$ | $1.74 \pm 1.57$ | $2.17 \pm 0.91$ | $1.31 \pm 0.73$ |


\end{document}